\documentclass[10pt,twocolumn,letterpaper]{article}

\usepackage{iccv}
\usepackage{color,soul}
\usepackage{array}
\usepackage{times}
\usepackage{epsfig}
\usepackage{graphicx}
\usepackage{caption}
\usepackage[T1]{fontenc}
\usepackage[utf8]{inputenc}
\usepackage[english]{babel}
\usepackage[labelformat=simple]{subfig}
 
\usepackage{amsmath}
\usepackage{amssymb}
\usepackage{blindtext, subfig}

\usepackage[breaklinks=true,bookmarks=false]{hyperref}

\iccvfinalcopy 

\def\iccvPaperID{****} 

\begin{document}

\title{Attended Temperature Scaling:\\ A Practical Approach for Calibrating Deep Neural Networks}


\author{ Azadeh Sadat Mozafari\\ 
{\tt\footnotesize azadeh-sadat.mozafari.1@ulaval.ca }
\and 
Hugo Siqueira Gomes\\
{\tt\footnotesize hugo.siqueira-gomes.1@ulaval.ca }\\
\and 
Wilson Leão\\
{\tt\footnotesize wilson.leao@petrobras.com.br }\\
\and 
Steeven Janny\\
{\tt\footnotesize steeven.janny@ens-paris-saclay.fr}
\and 
Christian Gagn\'e\\
{\tt\footnotesize christian.gagne@gel.ulaval.ca}
}

\maketitle

\begin{abstract}

Recently, Deep Neural Networks (DNNs) have been achieving impressive results on wide range of tasks. However, they suffer from being well-calibrated. In decision-making applications, such as autonomous driving or medical diagnosing, the confidence of deep networks plays an important role to bring the trust and reliability to the system. To calibrate the deep networks' confidence, many probabilistic and measure-based approaches are proposed. Temperature Scaling (TS) is a state-of-the-art among measure-based calibration methods which has low time and memory complexity as well as effectiveness. In this paper, we study TS and show it does not work properly when the validation set that TS uses for calibration has small size or contains noisy-labeled samples. TS also cannot calibrate highly accurate networks as well as non-highly accurate ones. Accordingly, we propose Attended Temperature Scaling (ATS) which preserves the advantages of TS while improves calibration in aforementioned challenging situations. We provide theoretical justifications for ATS and assess its effectiveness on wide range of deep models and datasets. We also compare the calibration results of TS and ATS on skin lesion detection application as a practical problem where well-calibrated system can play important role in making a decision.

\end{abstract}
\section{Introduction}
\label{introduction}


\begin{figure}[!ht]
 \centering
 
 \subfloat[][\scriptsize Label =  Dermatofibroma\\Pred. =  Dermatofibroma\\Confidence = 0.99\\Calib. Confidence = 0.98 ]{\label{fig:1random}\includegraphics[height=2.4cm,width=2.7cm]{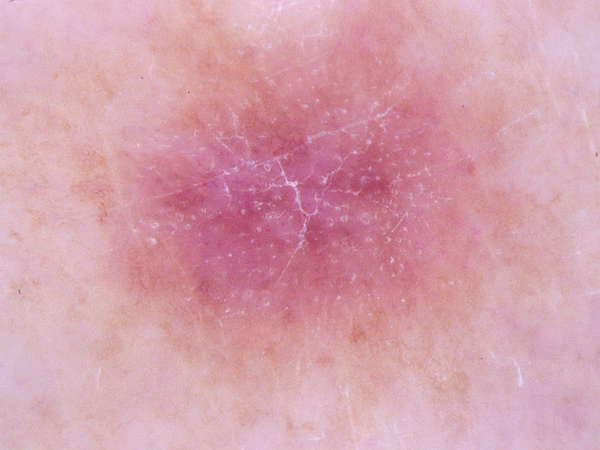}}
 \hfill
 \subfloat[][\scriptsize Label = BCC\\Pred. =  BCC\\Confidence = 0.99\\Calib. Confidence = 0.99 ]{\label{fig:2random}\includegraphics[height=2.4cm,width=2.7cm]{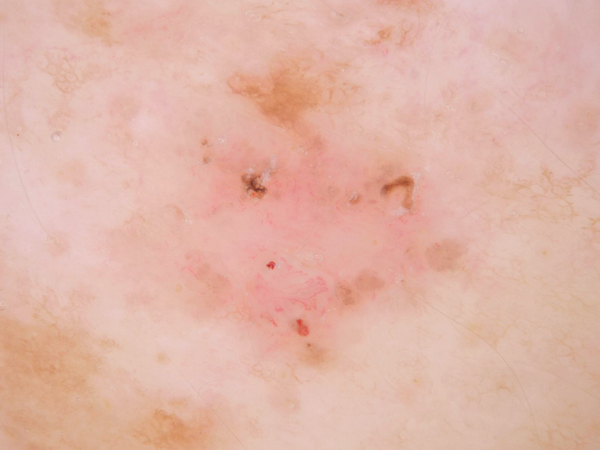}}
 \hfill
 \subfloat[][\scriptsize Label = Melanocytic nevus\\Pred. =  Melanocytic nevus\\Confidence = 0.99\\Calib. Confidence = 0.98 ]{\label{fig:3random}\includegraphics[height=2.4cm,width=2.7cm]{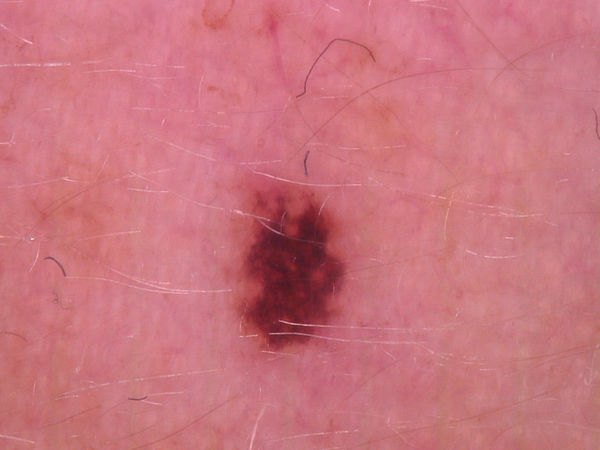}}\\
 
 \subfloat[][\scriptsize Label = Melanoma\\Pred. =  Bowen\\Confidence = 0.91\\Calib. Confidence = 0.54 ]{\label{fig:4random}\includegraphics[height=2.4cm,width=2.7cm]{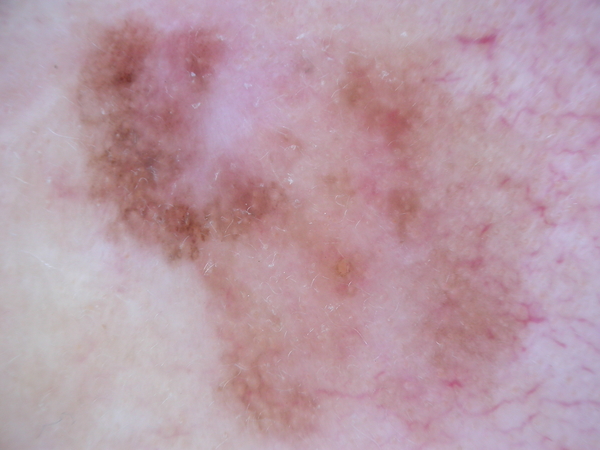}}
  \hfill
\subfloat[][\scriptsize Label = Bowen\\Pred. =  BCC\\Confidence = 0.90\\Calib. Confidence = 0.46 ]{\label{fig:5random}\includegraphics[height=2.4cm,width=2.7cm]{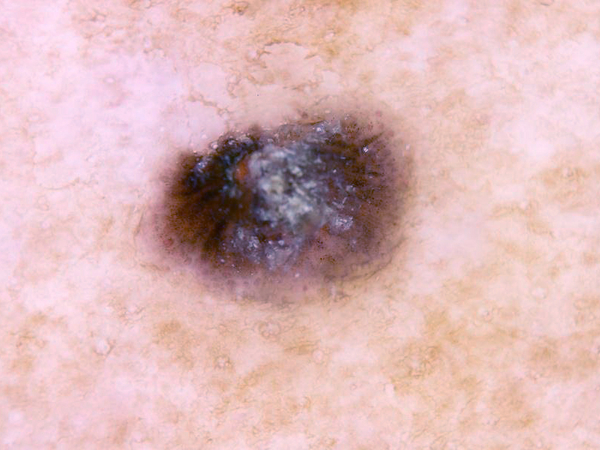}}
  \hfill
\subfloat[][\scriptsize Label = Benign keratosis\\Pred. =  Bowen\\Confidence = 0.89\\Calib. Confidence = 0.48 ]{\label{fig:6random}\includegraphics[height=2.4cm,width=2.7cm]{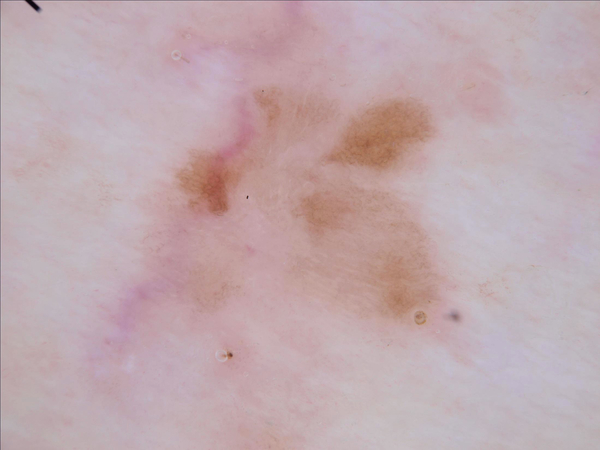}} 
\caption{Output of the medical assistant system for skin anomaly detection (more details in Section \ref{application}). Before calibration, the confidence of the system is high for both correctly and misclassified samples. After applying calibration, the network keeps the confidence of correctly classified samples high while decrease the confidence of misclassified samples.}
\label{fig_calibration}
\end{figure}

Deep Neural Networks (DNNs) show dramatically accurate results on challenging tasks such as computer vision \cite{he2016deep,simonyan2014very} speech recognition \cite{graves2013speech} and medical diagnosis \cite{bar2015chest}. However, in real-world decision-making applications, accuracy is not the only element considered and the confidence of the network is also essential for having a secure and reliable system. In DNNs, confidence usually corresponds to the output of a softmax layer, which is typically interpreted as the likeliness (probability) of different class occurrence. Most of the time, this value is far from the true probability of each class occurrence, with a tendency to get overconfident (i.e., output of one class close to 1 and other classes close to 0). In such case, we usually consider the DNN not to be well-calibrated.
Calibration in DNNs is a recent challenge in machine learning community  which was not an issue previously for shallow neural networks \cite{niculescu2005predicting}.
Gua et al. \cite{guo2017calibration} studies the role of different parameters which makes a neural network uncalibrated. They show a deep network which finds the optimal weights by minimizing Negative Log Likelihood (NLL) \cite{friedman2001elements} loss function, can reach to the higher accuracy when it gets overfitted to NLL. However, the side effect of overfitting to NLL is to make the network overconfident. 

Having calibrated network is important for real-world applications. In a self-driving car \cite{bojarski2016end} deciding about transferring the control of the car to the human observer is taken regarding to the confidence of the detected objects. In medical care systems \cite{jiang2011calibrating}, the deadly diseases can be missed when they are wrongly detected as a non-problematic case with high confidence. Calibration adds more information to the system which consequences reliability. Figure \ref{fig_calibration} compares the output of an overconfident system and calibrated one for misclassified and correctly classified samples in a skin lesion detection system. The calibrated networks decrease the confidence in the case of wrongly  detected samples while preserves the confidence for most of correctly classified ones. Calibration methods for DNNs are widely investigated  in recent literature and can be categorized into two main directions: 1-probabilistic approaches 2-measure-based approaches. Probabilistic approaches generally include approximated Bayesian formalism \cite{ mackay1992bayesian,neal2012bayesian,louizos2016structured,blundell2015weight}. In practice, the quality of predictive uncertainty in Bayesian-based methods relies heavily on the accuracy of sampling approximation and correctly estimated prior distribution. Despite of significant achievements in distribution estimation, these approaches are complex and suffer from a significant computational burden, time- and memory complexity.

Comparatively, measure-based approaches are more practical. They are generally post-processing methods that do not need to retrain the network to make it calibrated. 
Temperature Scaling (TS) \cite{guo2017calibration} is the state-of-the-art measure-based approach that comparing to the others, achieves better calibration with minimum computational complexity (optimizing only one parameter $T$ to soften the softmax) which makes it the most appealing method in practice. It also preserves the accuracy rate of the network that can be degraded during the calibration phase. In TS, the best $T$ parameter is found by minimizing NLL loss with respecting to $T$ on validation set.  One big challenge in real scenarios to apply TS is gathering enough number of samples for validation set and labeling them by an expert to calibrate an already pre-trained model. Asking non-professional experts to label the samples for decreasing the expenses, may bring labeling noise to the validation set, especially in medical applications. Therefore being robust to the noise and size of validation set is a rising need in calibration applications. Despite of TS interesting results, when the DNN is highly accurate, or the validation set is small or contains noisy labels, TS cannot calibrate the DNN successfully.    

\textbf{Contribution:}
In this paper, we propose a new TS family method which is called Attended Temperature Scaling (ATS) to make a better adjustment of confidence in DNNs. Comparing to TS algorithm, ATS preserves the time and memory complexity advantage of classic TS as well as intact accuracy while it brings better calibration. It specially works properly in the case of small-size validation set, highly accurate DNNs and validation set with labeling noise in which TS is not functioning well. We analyze theoretically ATS and demonstrate why it works better in these situations. 


\section{Related Works}
\label{Related Works}
Recently, in the literature there is interest in adapting NNs to encompass uncertainty. The studies are summarized into two categories of probabilistic and measure-based approaches. The probabilistic approach is referred to Bayesian theory \cite{bernardo2009bayesian} for estimating the conditional distribution of data. In these methods a prior distribution is defined on the parameter of a NN and given the training set, the posterior of the parameters will be computed.  This is the general definition of Bayesian Neural Network which brings back the uncertainty to DNNs framework. As the exact Bayesian inference is not practical, a variety of approximation are proposed such as Laplace approximation \cite{mackay1992bayesian, ritter2018scalable,ritter2018online,kirkpatrick2017overcoming}, Variational Bayesian methods \cite{molchanov2017variational,louizos2017multiplicative,blundell2015weight,louizos2016structured} and Monte Carlo Markov Chains (MCMC) \cite{neal2012bayesian,balan2015bayesian,chen2014stochastic} to make Bayesian deep networks tractable.
MC-dropout \cite{gal2016dropout} is another probabilistic approach which removes the complicated training setup of Bayesian models and replace it with simple dropout in training and test phases. Gal et al.~\cite{gal2016dropout} show MC-dropout approximates Variational Bayesian inference. The testing and training time complexity of MC-dropout is high. Although it is a simple approach to apply. Ensemble of DNNs \cite{lakshminarayanan2017simple} is another non-complicated probabilistic approach that can achieve better calibrated results than MC-dropout. In this method, an ensemble of different deep models are constructed by bagging. This approach is appropriate for the parallel computing with GPUs that can train multiple DNNs in the same time. However, keeping the models in the memory during the test time brings high memory complexity to this method.

Measure-based approaches are much less complex in applying calibration comparing to probabilistic approaches. In measure-based approach, the main idea is to decrease the miscalibration of the network by minimizing a loss which is a calibration measure. The common calibration measures are: Negative Log Likelihood (NLL), Expected Calibration Error (ECE) \cite{naeini2015obtaining} and Brier score \cite{brier1950verification}. Generally for training the neural network, NLL is used which simultaneously increases accuracy and decreases miscalibration. However, it easily gets overfitted and makes the network overconfident \cite{guo2017calibration}. Kumar et al \cite{kumar2018trainable} propose a RKHS kernel based measure which they call it MMCE as a derivable surrogate of ECE (ECE is not smooth function). They use MMCE with NLL as the loss function which is minimized during the training to get the network calibrated. The other group of measure-based approaches like Temperature Scaling \cite{ guo2017calibration}, Platt-Scaling \cite{ platt1999probabilistic} , Histogram Binning \cite{zadrozny2001obtaining}, Isotonic Regression \cite{zadrozny2002transforming} and Baysian Binning into Quantiles \cite{naeini2015obtaining} fine-tunes the softmax layer by keeping the DNNs' weights unchanged. They do not need to retrain the deep network from scratch and they only need to find the best parameter of softmax softening function  by minimizing a calibration loss on a small validation set. Therefore they are appropriate for real scenarios in which the time and memory complexity of calibration is the concern such as autonomous driving \cite{neumann2018relaxed} and weather forecasting \cite{kuleshov2018accurate}. 
In this paper, we have focused on Temperature Scaling family \cite{guo2017calibration} and propose a new TS approach called ATS that can come into better calibration.
\section{Problem Setup}
\label{Problem Setting}
In this section, we set up the problem, notations and introduce two different calibration measures that we will use in this paper.\\
\textbf{Assumptions}: We assume to have access to a pre-trained deep model $D(\cdot)$ with the ability of detecting $K$ different classes. $D(\cdot)$ is trained on samples generated from distribution function $Q(\mathbf{x},y)$. We also have access to a small validation dataset $\mathcal{V}= \{(\mathbf{x}_i,y_i)\}_{i=1}^N$ with the same distribution as the training and test set. 
For each sample $\mathbf{x}_i$, there exist $\mathbf{h}_i=[h_i^1,h_i^2,\ldots,h_i^K]^\top$ which is the logit layer. $D(\mathbf{x}_i) = (\hat{y_i},S_{y = \hat{y_i}}(\mathbf{x}_i))$ defines that the network $D(\cdot)$ detects label $\hat{y_i}$ for input data $\mathbf{x}_i$ and confidence $S_{y=\hat{y_i}}(\mathbf{x}_i)$.  $S_y(\mathbf{x})=exp({h_i^{{y}}})/ \sum_{j=1}^K exp({h_i^j})$ is the softmax output function of the model that here is interpreted as the confidence.\\
\textbf{Goal}: The objective is to adjust $\mathbf{h}_i$ with rescaling parameter $T$ in order to minimize calibration error of the model.



\subsection{ Measures for Calibration }
\label{measure for calibration}
Based on the different definitions given for calibrated model, two common measures are utilized in the literature which are NLL and ECE.

\subsubsection{ Negative Log Likelihood (NLL)~\cite{friedman2001elements}}
\label{NLL}
When the network is calibrated, the softmax output layer is supposed to have exact approximation of the true conditional distribution $Q(y|\mathbf{x})$. To measure the calibration, the amount of similarity between $S_y(\mathbf{x})$ and $Q(y|\mathbf{x})$ functions can be computed. As $Q(y|\mathbf{x})$ distribution function is not available and only some generated samples from it are available (validation set), the similarity can be computed based on Gibbs inequality given in Eq.~(\ref{eq1}):
\begin{equation}
-\mathbb{E}_{Q(\mathbf{x},y)}[\log \left( Q(y|\mathbf{x})\right)] \leq -\mathbb{E}_{Q(\mathbf{x},y)}[\log \left(P(y|\mathbf{x})\right)],
\label{eq1}
\end{equation}
where $\mathbb{E}$ is the expected value function. 
 The minimum of $-\mathbb{E}_{Q(\mathbf{x},y)}[\log P(y|\mathbf{x})]$ happens when $P(y|\mathbf{x})$ is equal to the true conditional distribution of $Q(y|\mathbf{x})$. This inequality is valid for any arbitrary distribution function $P(y|\mathbf{x})$. NLL is defined as the empirical estimation of $-\mathbb{E}_{Q(\mathbf{x},y)}[\log P(y|\mathbf{x})]$ which in deep neural networks is rephrased as:
\begin{equation}
\text{NLL} = -\sum_{(\mathbf{x}_i,y_i)}\log \left(S_{y=y_i}(\mathbf{x}_i)\right), \quad  (\mathbf{x}_i,y_i)\sim Q(\mathbf{x},y).
\label{eq2}
\end{equation}
NLL can be used as a calibration measure that shows the similarity between a probability function $S_y(\mathbf{x})$ and the true conditional distribution $Q(y|\mathbf{x})$ of data in which the smaller means more calibrated.

\subsubsection{ Expected Calibration Error (ECE)~\cite{naeini2015obtaining}}
Another way to define calibration is based on the relation between the accuracy and confidence. Miscalibration can be interpreted as the difference between confidence and probability of correctly classifying a sample. For instance, in the case of a calibrated model, if we have the group of samples which has the confidence of $S_{y}(\mathbf{x})=0.9$, it is supposed to have $0.9$ percentage of accuracy. Based on this definition of calibration, ECE is proposed as empirical expectation error between the accuracy and confidence. 
It is calculated by partitioning the range of confidence between $[0\,,1]$ into $L$ equally-spaced confidence bins and then assign the samples to each bin $B_l$ where $l=\{1,\ldots,L\}$ by their confidence range. Later it calculates the weighted absolute difference between the accuracy and confidence for each subset $B_l$. More specifically:
\begin{equation}
\text{ECE} = \sum_{l=1}^L{\frac{|B_l|}{N}}\Big|\text{acc}(B_l)-\text{conf}(B_l)\Big|,
    \label{eq4}
\end{equation}\\
where $N$ is the total number of samples. ECE is not derivable function, therefore we focus on NLL loss function as the measure for the proposed calibration method. However, we report the calibration error on NLL and ECE to show the model will get calibrated by both definitions.

\section{Temperature Scaling (TS)~\cite{guo2017calibration}}
\label{Temperature Scaling}

\begin{figure*}[ht]
\centering
 
\includegraphics[width=15cm, height = 5.5cm]{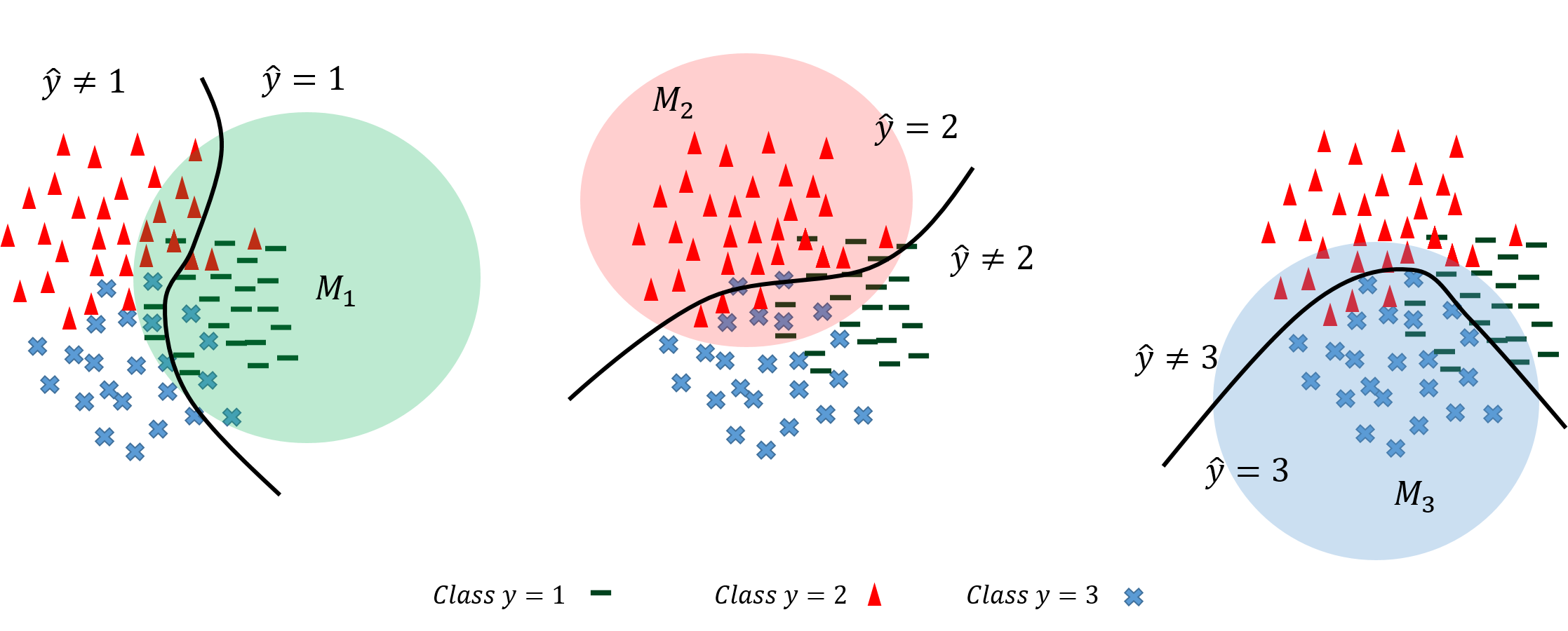}

\caption{A three class classification problem. The samples which are selected for each subset $M_k$ is shown inside the colored region. ATS considers two groups of samples to construct $M_k$ subset: the samples with true label $y=k$ and the samples with true label $y\not=k$ that located near to the decision boundary of that class (the black line).} 

\label{figure_M_division}
\end{figure*}

TS is a post-processing approach which rescales the logit layer of a deep model by parameter $T$ that is called temperature. TS is used to soften the output of the softmax layer and  makes it more calibrated. The best value of $T$ will be obtained by minimizing NLL loss function respecting to $T$ conditioned by $T>0$ on validation set $\mathcal{V}$ as defined in Eq.~(\ref{Eq(5)}):
\begin{equation}
\begin{split}
\label{Eq(5)}
    T^* =&\operatorname*{arg\,min}_{T} \left(-\sum_{i=1}^N\log\big(S_{y = y_i}(\mathbf{x}_i,T)\big)\right)\hspace {3mm}\\    &S.t: T>0,  \hspace {3mm} (\mathbf{x}_i,y_i)\in \mathcal{V},
\end{split}
\end{equation}
where $S_{y=y_i}(\mathbf{x}_i,T)= exp({\frac{h_i^{y_i}}{T}})/ \sum_{j=1}^K exp(\frac{h_i^j}{T})$, is the softed version of softmax by applying parameter $T$. TS has the minimum time and memory complexity among calibration approaches as it only optimizes one parameter $T$ on small validation set. Having only one parameter helps TS not only to be efficient and practical but also not to get overfitted to NLL loss function when it is optimized on small validation set $\mathcal{V}$. 

\subsection{Analyzing TS Approach}
\label{disscusion TS}
TS previously is applied for calibration \cite{guo2017calibration}, distilling the knowledge \cite{hinton2015distilling} and enhancing the output of DNNs for better discrimination between the in and out distribution samples \cite{liang2017enhancing}. It only rescales the output of logit layer to calibrate the network which causes preserving the accuracy unchanged. For the post-processing calibration approaches, keeping the accuracy intact is an important property as the other calibration methods which can change the accuracy is in danger of overfitting to the validation set and accuracy drop. Referring to Eq.~(\ref{Eq(5)}), by computing the derivative of NLL respecting to $T$ and putting it equal to zero, we will have :
\begin{equation}
\label{TS derivitive}
    \sum_{i=1}^Nh_i^{y_i} = \sum_{i=1}^N\sum_{k=1}^Kh_i^kS_{y=k}(\mathbf{x}_i,T^*).
\end{equation}
It shows regarding to the true label of the samples, TS selects the $T$ value which maximizes the $S_{y=k}(\mathbf{x}_i,T^*)$  for $k=y_i$ and minimize $S_{y=k}(\mathbf{x}_i,T^*)$ for all the other $k\not=y_i$. Therefore for correctly classified samples that $y_i = \arg\max(S_{y}(\mathbf{x}_i))$ the maximum confidence of the DNN is for the true class of the samples, $T$ approaches $0$ to increase the confidence twoard $1$. For missclassified samples, $T$ goes toward $\infty$ to decrease the confidence of predicted label which is misclassified and increase the confidence of the true label toward $1/K$. The balance between the correctly classified and misclassified samples brings back the optimal point $T$. 
When the validation set does not contain enough correctly and misclassified samples, TS finds the suboptimal $T$ value. This case happens in calibrating the highly accurate classifiers that the number of misclassified samples for them is few or when the size of validation set is small.  TS is also sensitive to the noise of the labels as the optimal $T$ value is dependent strongly on the true label of the samples in NLL loss function.  
\section{Attended Temperature Scaling (ATS)}
\label{ATS}
\begin{table*}[ht!]
\scriptsize
\centering
\caption{The results of different measure-based calibration methods for variation of datasets and models. ATS achieves the best calibration results for almost all experiments. }
\resizebox{\textwidth}{!}{  
\begin{tabular}{|l|l|>{\centering}m{1cm}|c|c|c|c|c|c|c|c|c|c|c|c|}
\hline
  
               &                              &\textbf{Uncalib. TS, ATS} &\multicolumn{2}{c|}{\textbf{Uncalibrated}}  & \multicolumn{2}{c|} {\textbf{TS}}          &\multicolumn{2}{c|} {\textbf{ATS}}   & \multicolumn{3}{c|} {\textbf{Matrix Scaling}}       & \multicolumn{3}{c|} {\textbf{Vector Scaling}}              \\
\textbf{Model} & \textbf{Dataset} &ACC &NLL  &ECE\%   &NLL   &ECE\%   &NLL   &ECE\%   &ACC &NLL &ECE\%    &ACC &NLL &ECE\% \\
\hline

VGG16 & Birds &75.975\% &0.929 &6.033 &0.929 &6.021 &\textbf{0.919} &\textbf{3.572} &74.870\% &0.961 & 5.683 &\textbf{76.99\%} &1.153 &11.992\\
ResNet152  & ImageNet  &76.71\% &0.935 &5.935 &0.927 &5.412 &\textbf{0.900} &\textbf{1.982} &76.03\% &0.935 &5.932 &\textbf{76.99\%} &1.175 &12.214\\

DenseNet40 & CIFAR10     &\textbf{92.61\%} &0.286 &4.089      &0.234   &3.241  &\textbf{0.221} &\textbf{0.657} &81.50\% &0.590   &5.793     &92.09\% &0.360 &4.939 \\

DenseNet40 & CIFAR100    &\textbf{71.73\%}  &1.088 &8.456       &\textbf{1.000}   &1.148  &\textbf{1.000} &\textbf{1.004} &57.50\% &1.918 &19.269     &32.48\% &9.655 &52.609\\

DenseNet100 & CIFAR10    &\textbf{95.06\%} &0.199 &2.618       &\textbf{0.156} &0.594 &\textbf{0.156} &\textbf{0.580}  &94.38\% &0.191   &2.272  &94.97\%  &0.247 &3.263 \\

DenseNet100 & CIFAR100   &\textbf{76.21\%} &1.119  &11.969    &0.886 &4.742  &\textbf{0.871} & \textbf{1.583}  &63.92\% &1.857 &19.977 &73.67\%  &1.602 &18.073\\

DenseNet100 & SVHN  &95.72\% &0.181  &1.630      &0.162 &0.548     &\textbf{0.161}   &\textbf{0.514} &95.73\%  &0.170 &1.126   &\textbf{95.99\%}  &0.162  &0.548  \\

ResNet110 & CIFAR10  &\textbf{93.71\%}  & 0.312  &4.343     &0.228  &4.298  &\textbf{0.206}  &\textbf{0.972} &92.13\% &0.285  &3.597   &93.17\% &0.375  &5.033   \\

ResNet110 & CIFAR100  &\textbf{70.31\%}  & 1.248   &12.752       &1.051 &1.804    &\textbf{1.050} &\textbf{1.529} &58.03\% &2.074  &20.749   &68.15\% &1.705 &19.751\\

ResNet110 & SVHN   &\textbf{96.06\%} &0.209  &2.697   & 0.158 &1.552 &\textbf{0.154} & \textbf{0.849} &96.00\% &0.173 &1.769   &\textbf{ 96.06\%} &0.254 &2.946\\

WideResNet32 & CIFAR100    &\textbf{75.41\%} & 1.166  &13.406     & 0.909 & 4.096   & \textbf{0.891} & \textbf{2.511} &66.18\%  &1.673  &18.265   &73.17\% & 1.693  & 18.7   \\

LeNet 5  &MNIST &\textbf{99.03\%} &0.105 &0.727    &0.061 &0.674    &\textbf{0.0341} &\textbf{0.354} &98.59\% &0.115 &1.152   &98.33 \% &0.048  &0.668 \\

VGG16 & CIFAR10 &\textbf{92.09\%} &0.427 &5.99  & 0.301 &6.015 &\textbf{0.271} &\textbf{2.978} &90.66\%  &0.364 &5.776 &91.90\% &0.432 &5.968\\

VGG16 & CIFAR100 &\textbf{69.00\%} &1.984 &21.493  &1.283 &\textbf{8.072} &\textbf{1.273} &8.283 &51.30\%  &2.290 &19.022  &68.92\% &2.441 &23.055\\

\hline
\end{tabular}
}

\label{Tabel_calibration}
\end{table*}

\begin{table}[ht!]
\scriptsize
\centering
\caption{The selected threshold $\theta$, TS temperature and ATS temperature for the datasets and models reported in Table \ref{Tabel_calibration}. }
\begin{tabular}{|p{1.8cm}|p{1.3cm}|p{0.8cm}|p{0.8cm}|p{0.8cm}|}
\hline
               
\textbf{Model} & \textbf{Dataset} &\textbf{$T_{TS}$}  &\textbf{$T_{ATS}$}  &{$\theta$} \\
& & & &\\
\hline
VGG16 & Birds  &1.000 &1.101 &0.01\\

ResNet152 & ImageNet  &1.064  &1.322  &0.03 \\ 

DenseNet40 & CIFAR10 & 2.505 & 1.921 & 0.04 \\
DenseNet40 & CIFAR100 & 1.450 & 1.435 & 0.02 \\

DenseNet100 & CIFAR10 & 1.801 &1.809 & 0.05\\

DenseNet100 & CIFAR100 & 2.178  & 1.904 & 0.01 \\

DenseNet100 & SVHN & 1.407   & 1.355   &0.05\\ 

ResNet110 & CIFAR10 & 2.960 & 2.321 & 0.02 \\

ResNet110 & CIFAR100 & 1.801 & 1.630 & 0.02 \\

ResNet110 & SVHN & 2.090 & 1.975  & 0.01\\ 

WideResNet32 & CIFAR100 & 2.104 & 1.971 & 0.01  \\

LeNet 5 &MNIST & 1.645 &2.832  &0.0008  \\
VGG16 & CIFAR10  &3.229 &1.941  &0.02  \\
VGG16 & CIFAR100  &2.741 &1.968  &0.01  \\

\hline
\end{tabular}
\label{Tabel_Threshold}
\end{table}

TS cannot find the optimal $T$ value when the number of samples in validation set is not enough. The idea of ATS is  increasing the number of samples in the validation set with low computational cost. Previously, TS minimizes NLL to decrease the dissimilarity between the $S_y(\mathbf{x},T)$ and $Q(y|\mathbf{x})$. Instead, ATS attends to the conditional distribution of each class and decreases the dissimilarity between $S_{y=k}(\mathbf{x},T)$ and $Q(y=k|\mathbf{x})$ for each class $k=1,\ldots,K$. This setting brings a chance to increase the number of samples and  robustness to the noise. As the first step, ATS should gather the samples from each class  distribution  $Q(\mathbf{x},y=k)$. It divides $\mathcal{V}$ into $K$ sub validation sets $M_k$ which are supposed to contain the samples generated from the $Q(\mathbf{x},y=k)$. Naturally, the samples whose true label is $y=k$ belongs to this subset. ATS also adds more samples from $y\not=k$ classes (samples with different class label than $k$) to $M_k$ which have high probability to be generated from distribution $Q(\mathbf{x},y=k)$. 
Selecting the most probable samples from other classes is done based on:
\begin{equation}
\label{Eq(6)}
    Q(\mathbf{x},y=k) = \frac{Q(y=k|\mathbf{x})}{Q(y\neq k|\mathbf{x})}Q(\mathbf{x},y \neq k)
\end{equation}
 which is derived from Bayesian Theorem \cite{jin2017introspective}. Eq.~(\ref{Eq(6)}) says the probability that a sample belongs to distribution $Q(\mathbf{x},y\neq k)$ is equal to the probability of belonging to the distribution $Q(\mathbf{x},y=k)$ by applying weight $W = Q(y=k|\mathbf{x})/Q(y\neq k|\mathbf{x})$.
 Notice that in this equation, $Q(y\neq k|\mathbf{x})$ is equal to $1-Q(y=k|\mathbf{x})$. 

Referring to Eq.~(\ref{Eq(6)}), ATS selects the samples with true label $y\not=k$ which are more probable to belong to  $Q(\mathbf{x},y=k)$, i.e. selecting the samples with bigger $W$. To calculate $W$, ATS considers $S_{y=k}(\mathbf{x})$ as an approximation to $Q(y=k|\mathbf{x})$. It selects the samples whose $S_{y=k}(\mathbf{x})$ is bigger than a threshold $\theta$. Those samples are generally located near to the boundary of classifier which separates class $k$ from class $\not=k$. Figure \ref{figure_M_division} gives an example of selected samples for $M_k$ in the case of three class classification problem. Formal saying $M_k$ is:

\begin{table}[ht!]
\scriptsize
\centering
\caption{The calibration results of probabilistic-based approaches MC-dropout and ensemble.  }
{  
\begin{tabular}{|p{0.8cm}|p{1cm}|p{0.6cm}|p{0.5cm}|p{0.5cm}|p{0.6cm}|p{0.5cm}|p{0.5cm}|}

\hline
               
    &                                         & \multicolumn{3}{c|} {\textbf{MC-Dropout}}       & \multicolumn{3}{c|} {\textbf{Ensemble}}              \\
\textbf{Model} & \textbf{Dataset} &ACC &NLL  &ECE\%   &ACC &NLL   &ECE\% \\
\hline

LeNet 5 & MNIST  & 99.33\% &0.044 & 0.453 &99.12\% &0.026 &0.307   \\
VGG16  & CIFAR10  &93.32\% &0.262 &4.239 &92.1\% & 0.342 & 5.99 \\
VGG16 & CIFAR100  & 70.45\% &1.527 &18.25 & 74.38\% &1.218 &6.756 \\
\hline

\end{tabular}
}
\label{Tabel_mcdropout}
\end{table}

\begin{equation}
\label{Eq(7)}
   M_k=\{(\mathbf{x}_i,y_i) \hspace {2mm}|\hspace {2mm} y_i = k \hspace {2mm} or\hspace {2mm} S_{y=k}(\mathbf{x}_i)\geq \theta\}
\end{equation}

threshold $\theta$ is a hyperparameter that will be fine-tuned on the validation set. After preparing the samples generated from each class distribution $Q(\mathbf{x},y=k)$, the optimal $T$ value will be found by minimizing the dissimilarity between $S_{y=k}(\mathbf{x},T)$ and $Q(y=k|\mathbf{x})$. As ATS changes the distribution of data by adding surrogate samples from other classes to make $M_k$ subsets, NLL loss function is not a calibration measure for it anymore. Therefore, we propose a new calibration measure for ATS which can be used as the loss function to find the optimal $T$:

\begin{equation}
\begin{split}
\label{Eq(8)}
   &\mathcal{L}_{ATS} = \sum_{k=1}^{K}\sum_{(\mathbf{x}_i,y_i)\in M_k}-\log\Big(\frac{S_{y=k}(\mathbf{x}_i,T)(1-S_{y = y_i}(\mathbf{x}_i,T))}{S_{y\not=k}(\mathbf{x}_i,T)}\Big),\hspace {3mm}\\
   & \quad \quad T^* = \operatorname*{arg\,min}_{T}(\mathcal{L}_{ATS}),
   \quad \quad S.t: T>0,
\end{split}
\end{equation}\\
Eq.~(\ref{Eq(8)}) describe the problem as the binary calibration setting where the labels of the samples are  $y_i\in\{k, \not=k\}$ for each subset $M_k$. We show in supplementary materials that minimizing $\mathcal{L}_{ATS}$ with respecting to $T$ on $M_k$ leads $S_{y=k}(\mathbf{x},T^*)$ to approach $Q(y=k|\mathbf{x})$ for all $k=1,\ldots,K$. Consequently,  $S_{y}(\mathbf{x},T^*)$ approaches $Q(y|\mathbf{x})$ which means the model get calibrated. In other words, $\mathcal{L}_{ATS}$ is a calibration measure for ATS that by minimizing it the model will get calibrated.\\






\subsection{Analyzing ATS Approach}
\label{Analyzing ATS}
ATS considers $M_k$ as the samples which are generated from the $Q(\mathbf{x},y=k)$ distribution regardless of their true labels. Therefore it is more robust to the noise than TS. Even for the term $(1-S_{y = y_i}(\mathbf{x}_i,T))$ in $\mathcal{L}_{ATS}$ which needs the true label of the sample, because it changes the problem to two-class-label $y_i\in\{k, \not=k\}$ instead $y_i \in\{1,\ldots,K\}$, it decreases the influence of the noise. ATS also increases the number of samples for estimating $Q(y=k|\mathbf{x})$ by reusing the other class samples in the validation set which leads finding more accurate $T$ values for small validation set. As the added samples are mostly near to the decision boundary, ATS also can improve calibration for the almost accurate DNNs that have few number of misclassified samples to cover that part of distribution.     

\section{Experiments}
\label{Experiments}

We conduct the experiments in two parts: 1- analyzing behavior of ATS 2- Impact of better calibration in making a decision maker application more reliable. In the first part, we compare ATS with different measure-based and probabilistic-based post-processing approaches in calibrating a wide range of model-dataset. We also compare the robustness of ATS versus TS to the noise of the labels and validation size to show empirically ATS is more reliable and robust than TS. In the second part, we demonstrate the impact of improvement in calibration for making a lesion diagnose system reliable. The list of models and datasets which are used in the experiments are as follows (more details are provided in supplementary material):\\
\textbf{Datasets}:
To investigate the validity of ATS, we test the methods on CIFAR-10 \cite{krizhevsky2009learning}, CIFAR-100 \cite{krizhevsky2009learning}, SVHN \cite{netzer2011reading}, MNIST \cite{lecun1998mnist}, Calthec-UCSD Birds \cite{wah2011caltech} and ImageNet2012 \cite{deng2009imagenet}. For the medical application we use data extracted from the “ISIC 2018: Skin Lesion Analysis Towards Melanoma Detection” grand challenge datasets \cite{DBLP:journals/corr/abs-1710-05006,tschandl2018ham10000} that contains images of 7 different skin lesion types.\\
\textbf{Models}: We try wide range of different state-of-the-art deep convolutional networks with variations in depth. The selected DNNs are Resnet \cite{he2016deep}, WideResnet \cite{zagoruyko2016wide}, DenseNet \cite{iandola2014densenet}, Lenet \cite{lecun1998gradient}, and VGG \cite{simonyan2014very}. We use the data pre-processing, training procedures and hyper-parameters as described in each paper.\\
\textbf{Experiment setup}:
In all experiments, we select $20\%$ of the test set as validation and the rest as the test dataset to report the results on. For the experiments that need different-size validation sets, the data will be selected randomly from the validation set is already separated from the $20\%$ of the test.  For ATS method, hyper-parameter $\theta$ will be fine-tuned on validation to minimize NLL based on the returned $T$ value by ATS. According to the accuracy of the network, the search interval for $\theta$ is changing between $[0,1]$, $[0,0.1]$ or $[0,0.001]$ and the search step of it would change between $0.01$, $0.001$ or $0.0001$ respectively. When the model is more accurate or the dataset contains more classes the search interval becomes smaller with smaller steps. \\
\textbf{Baselines to Compare}:
We compared ATS with different post-processing calibration methods which are:
\begin{enumerate}
    \item \textit{Temperature Scaling} \cite{guo2017calibration}: It is explained in Sec. (\ref{Temperature Scaling})
    \item \textit{Matrix and Vector Scaling} \cite{platt1999probabilistic}: Matrix Scaling applies a linear transformation on the logits: 
    \begin{equation}
    \begin{split}
        &S_{y=\hat{y_i}}({x_i},\boldsymbol{W},\boldsymbol{b}) = \underset{k}{\text{max}} \ \sigma(\boldsymbol{W}.\boldsymbol{h_i}+\boldsymbol{b})^{(k)}\\
        & \hat{y_i} = \operatorname*{arg\,max}_{k} \ \sigma(\boldsymbol{W}.\boldsymbol{h_i}+\boldsymbol{b})^{(k)}\\
    \end{split}
    \end{equation}

     Where $\sigma$ is the softmax function which takes logit layer as an input, and $S_{y=\hat{y}}(\mathbf{x}_i,\mathbf{W},\mathbf{b})$ is the confidence of sample $(\mathbf{x_i},y_i)$. The parameters $\boldsymbol{W}_{K\times K}$ and $\boldsymbol{b}_{K}$ are optimized with respect to NLL on the validation set. Vector Scaling is the relaxed version of Matrix Scaling in which $\boldsymbol{W}_{K\times K}$ is a diagonal matrix.
    \item \textit{MC-dropout}  \cite{gal2016dropout}: A pre-trained model which is trained with dropout rate $p = 0.5$ is used with keeping the dropout active during the test. Each sample will be tested 100 times and the average of 100 confidences will be used as the final decision. 
    \item \textit{Ensemble} \cite{lakshminarayanan2017simple}: The ensemble of 3 same-architecture DNN models, which are trained with different random initial weights. The final confidence is the average of the confidence of the models.  
    
\end{enumerate}
\textbf{Evaluation metric} 
We report the results based on two calibration metric ECE and NLL which are already explained in Sec. (\ref{measure for calibration}). For ECE we set the number of bins to 15 for all the experiments.     

\begin{figure*}[!ht]
 \centering
 \subfloat{\label{fig:gull}\includegraphics[height=4cm,width=4cm]{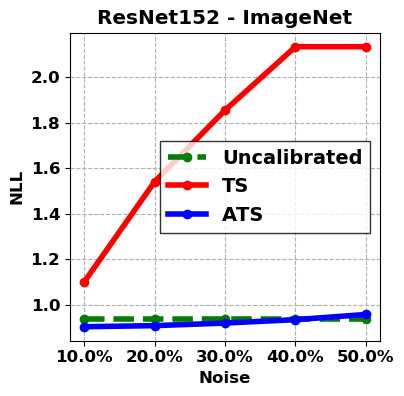}}
 \subfloat{\label{fig:gull}\includegraphics[height=4cm,width=4cm]{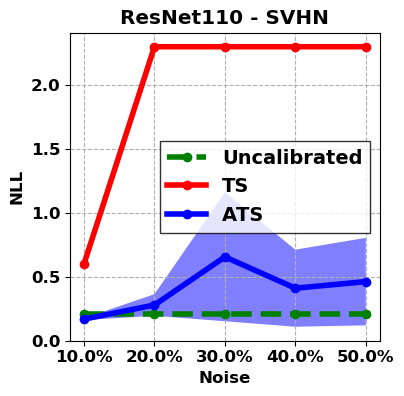}}
 \subfloat{\label{fig:gull}\includegraphics[height=4cm,width=4cm]{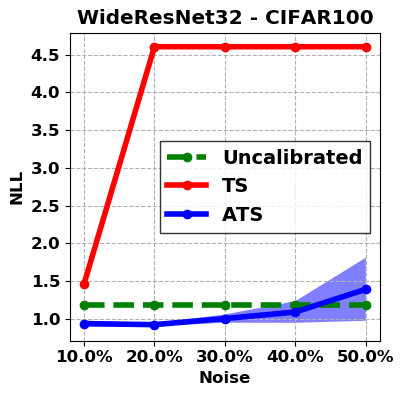}}
 \subfloat{\label{fig:gull}\includegraphics[height=4cm,width=4cm]{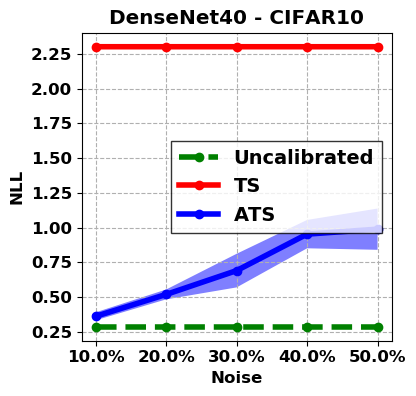}}
 \caption{ Calibration of different models-datasets with TS and ATS methods for $10\%\sim 50\%$ of labeling noise.  }
  \vspace{-0.3cm}
\label{noise}
\end{figure*}

\begin{figure*}[!ht]
 \centering
 \subfloat{\label{fig:gull}\includegraphics[height=4cm,width=4cm]{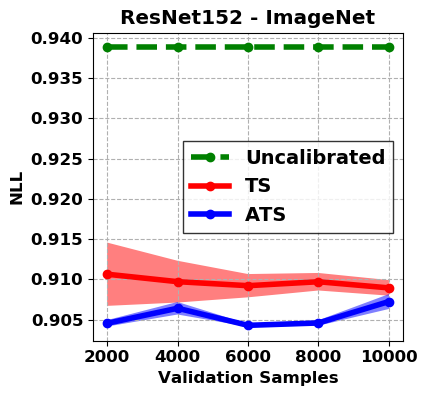}}
 \subfloat{\label{fig:gull}\includegraphics[height=4cm,width=4cm]{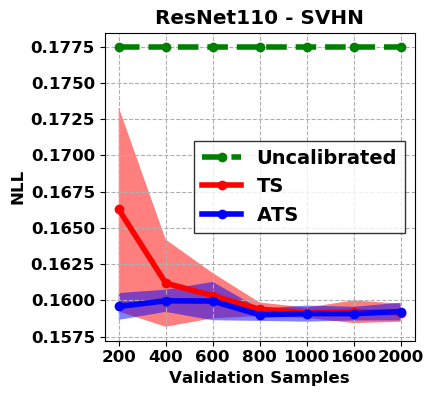}}
 \subfloat{\label{fig:gull}\includegraphics[height=4cm,width=4cm]{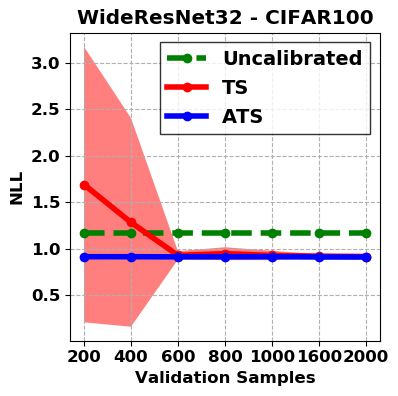}}
 \subfloat{\label{fig:gull}\includegraphics[height=4cm,width=4cm]{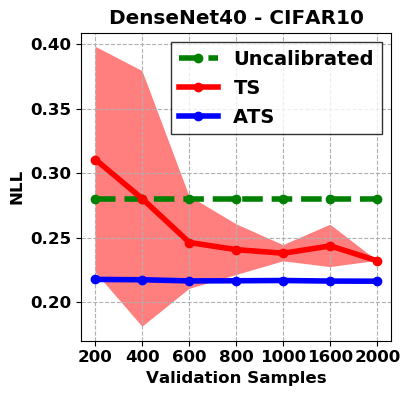}}
 \caption{ Calibration of different models-datasets with TS and ATS methods for different validation size.  }
\label{validation}
 
\end{figure*}

\subsection{Results}

\subsubsection{Calibration}
In Table \ref{Tabel_calibration}, the calibration result of TS and ATS which have only one parameter for fine-tuning the softmax output layer is compared with Matrix and Vector scaling which apply a linear function on logit layer. TS and ATS achieve better calibration results as well as preserving the accuracy rate of the network. It seems, however, Matrix and Vector Scaling can define more complex functions to soften the softmax layer, they suffer from over-fitting to the validation set in both accuracy and confidence. ATS in most cases calibrates the network better than all the others. Especially in the case of highly accurate networks such as VGG16-CIFAR10, ResNet110-SVHN, ResNet110-CIFAR10 and LeNet5-MNIST the calibration error improvement is more than the moderate accurate networks. This can be explained by the lack of enough misclassified samples in the validation set which prevent TS from converging to the local optima. Different $T$ values which are found for TS and ATS methods with the selected threshold $\theta$ are provided in Table \ref{Tabel_Threshold}. It shows even a small change in the value of $T$ can improve NLL significantly.  

In Table \ref{Tabel_mcdropout}, we provide the calibration result of two post-processing probabilistic approaches. To compare the results with other measure-based approaches, we also report the results of the same model-datasets in the last three lines of Table \ref{Tabel_calibration}.  The only method that uses dropout during the training is MC-Dropout. It shows MC-Dropout and ensemble can improve the accuracy and calibration of the model simultaneously and make the model more calibrated than other measure-based approaches. However, they increase time and memory complexity and they cannot be used to calibrate an already pre-trained models. 

\subsubsection{Robustness to Noise}
As explained in Sec. \ref{Analyzing ATS}, theoretically ATS is more robust to noise than TS. In this experiment, we empirically investigate the robustness of ATS to noise in comparing to TS method. We select four different combination of model-datasets and apply $5$ times a range of random noise from $10\%$ to $50\%$ to labels and report the mean and std of NLL. The behavior of ATS and TS are depicted in Figure \ref{noise} (more diagrams are provided in supplementary material). It shows TS is sensitive to the label noise and even with few percentage of labeling noise it cannot converge to optimal $T$. However, ATS is more robust to labeling noise and still can calibrate the model when the labels is not severely defected.

\subsubsection{Robustness to the Size of Validation Set }
ATS is much less sensitive to the size of the validation set comparing to TS. Figure \ref{validation} depicts the results for ATS vs. TS in calibrating four different model-datasets (more results are provided in supplementary material).  Each experiment is conducted $5$ times and the mean and std of NLL are reported. When the validation size is small, TS is not stable. By increasing the number of samples, TS converges to find the optimal $T$ value. ATS comparing to TS can find the optimal $T$ by smaller size of the validation set and it is more stable.

\begin{table*}[ht!]
\tiny
\centering
\caption{Comparing calibrating ResNet200 model trained on ISIC dataset with TS and ATS approaches. }
\resizebox{\textwidth}{!}{  
\begin{tabular}{|l|l|c|c|c|c|c|c|c|c|c|c|}
\hline
                                    \textbf{Model} &\textbf{Dataset}
                                    &\textbf{ACC}
                                                    &\multicolumn{2}{c|}{\textbf{Uncalibrated}}      &\multicolumn{2}{c|} {\textbf{TS}}          &\multicolumn{2}{c|} {\textbf{ATS}}  &\textbf{$T_{TS}$} &\textbf{$T_{ATS}$} &\textbf{$\theta$}  \\
       & & &NLL   &ECE\%   &NLL &ECE\%    &NLL &ECE\% &  &  &\\
\hline
ResNet200 &ISIC  &89.14\% &0.696 &8.374 &0.379 &7.978 &0.333 &1.157 &4.934 &3.548 &0.01\\
\hline
\end{tabular}
}
\vspace{-0.3cm}
\label{Tabel_calibration_ISIC}
\end{table*}

\subsection{Application}
\label{application}
In this section, we show the impact of calibration to improve the decision making in medical assistant system. One of medical applications is anomaly detection for skin spots.
We use ISIC dataset \cite{DBLP:journals/corr/abs-1710-05006,tschandl2018ham10000}, which contains 10015 images of 7 different skin lesion types. We divide the dataset into two randomly selected parts of training and test samples with 6009 and 4006 samples, respectively. As the number of samples for training is not enough we applied data augmentation which increases both the amount and diversity of data by randomly augmenting to train the ResNet200. Training settings and details of application is reported in supplementary materials. 

In a medical assistant system, the final confidence is used to refer the patient to specialist for further experiments. Therefore, the ideal detection system should be certain when it correctly classifies a sample and uncertain when it missclasifies a sample. However, it is obvious that the system cannot be certain about the correctly classified samples which are located near to the decision boundary. The ResNet200 trained model on the dataset is overconfident. We divide the test set into 801 images validation set and 3205 test images. Later, we apply TS and ATS to calibrate the model. The results of calibration is reported in Table \ref{Tabel_calibration_ISIC}. In a referral system, the samples that have higher confidence than a specific threshold, they will be accepted as the correctly classified and for the samples which have the confidence less than the threshold it will be referred to the specialist for further experiments. 

\begin{figure}[!ht]
 \centering
 \subfloat{\label{fig:gull}\includegraphics[height=5.4cm,width=7.7cm]{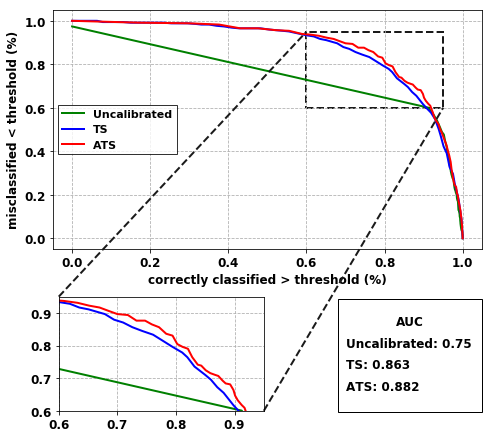}}
 \hfill
 
\caption{Reliability diagram of the skin lesions system. In the same rate of correctly classified samples upper than specific threshold, the system that have more misclassified samples lower than that threshold is more reliable.} 
\vspace{-0.3cm}
\label{fig_threshold}
 
\end{figure}

\begin{figure}[!ht]
 \centering
 \subfloat[][\scriptsize Label = Melanocytic nevus\\Pred. = Melanocytic nevus\\Confidence = 0.99\\ATS Confidence = 0.92\\TS Confidence = 0.58 ]{\label{fig:gull}\includegraphics[height=2.4cm,width=2.7cm]{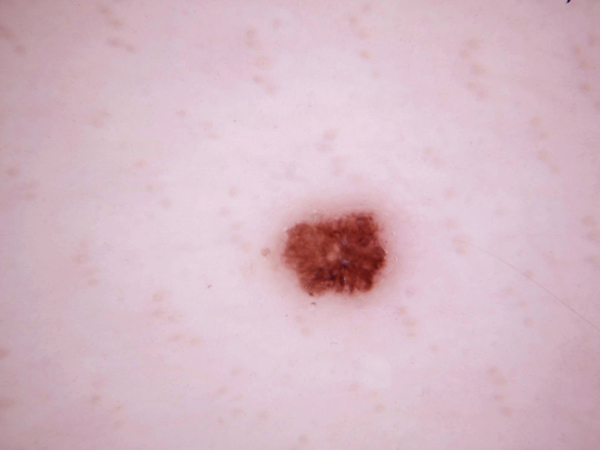}}
 \hfill
 \subfloat[][\scriptsize Label = Melanoma\\Pred. =  Melanoma\\Confidence = 0.99\\ATS Confidence = 0.93\\TS Confidence = 0.58 ]{\label{fig:gull}\includegraphics[height=2.4cm,width=2.7cm]{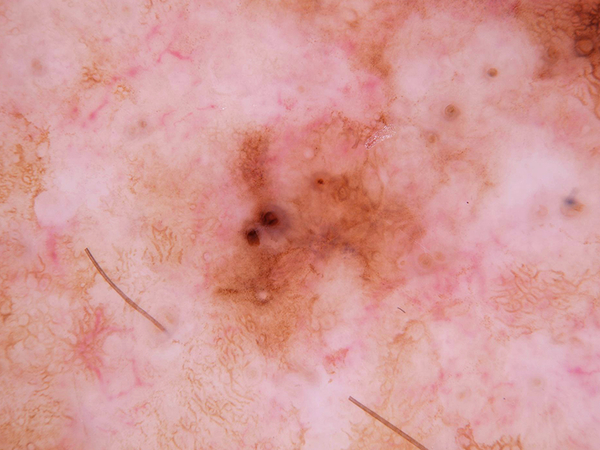}}
 \hfill
 \subfloat[][\scriptsize Label = BCC\\Pred. = BCC\\Confidence = 0.99\\ATS Confidence = 0.92\\TS Confidence = 0.58 ]{\label{fig:gull}\includegraphics[height=2.4cm,width=2.7cm]{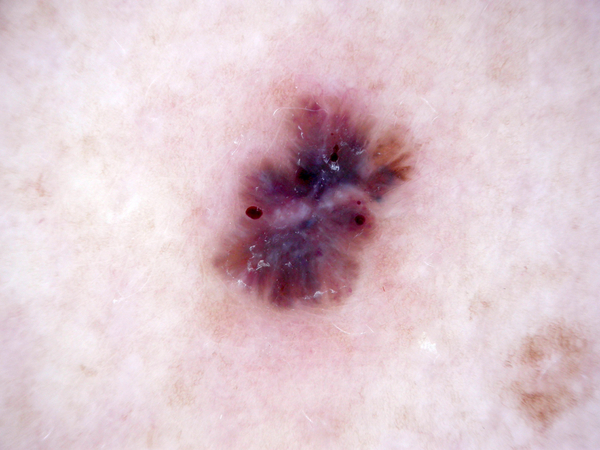}}\\

 \subfloat[][\scriptsize Label = Benign keratosis\\Pred. =  Melanoma\\Confidence = 0.92\\ATS Confidence = 0.65\\ TS Confidence = 0.41 ]{\label{fig:gull}\includegraphics[height=2.4cm,width=2.7cm]{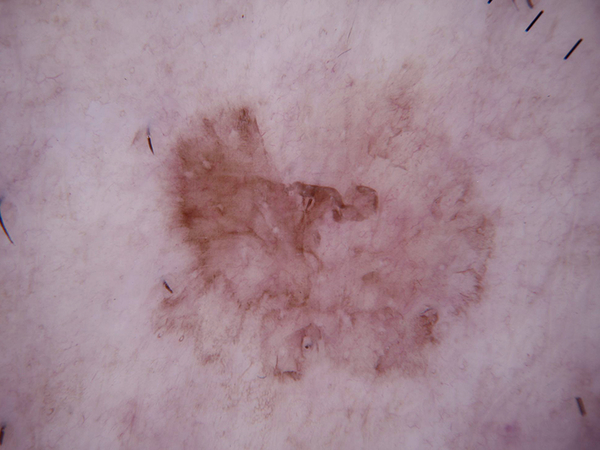}}
  \hfill   
\subfloat[][\scriptsize Label = Melanoma\\Pred. =  Melanocytic nevus\\Confidence = 0.93\\ATS Confidence = 0.68\\TS Confidence = 0.43 ]{\label{fig:gull}\includegraphics[height=2.4cm,width=2.7cm]{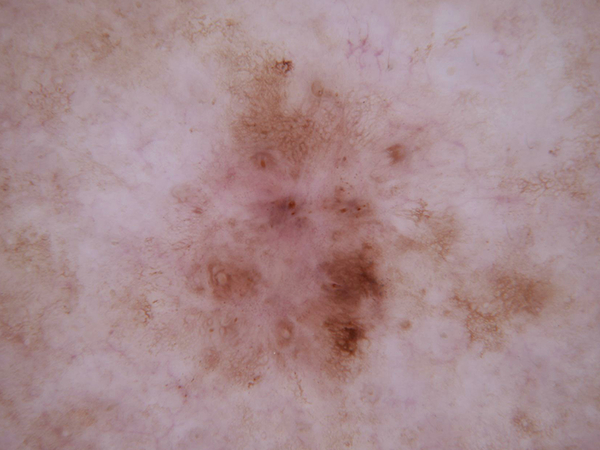}}
  \hfill
\subfloat[][\scriptsize Label = Bowen\\Pred. =  Benign keratosis\\Confidence = 0.87\\ATS Confidence = 0.65\\TS Confidence = 0.45 ]{\label{fig:gull}\includegraphics[height=2.4cm,width=2.7cm]{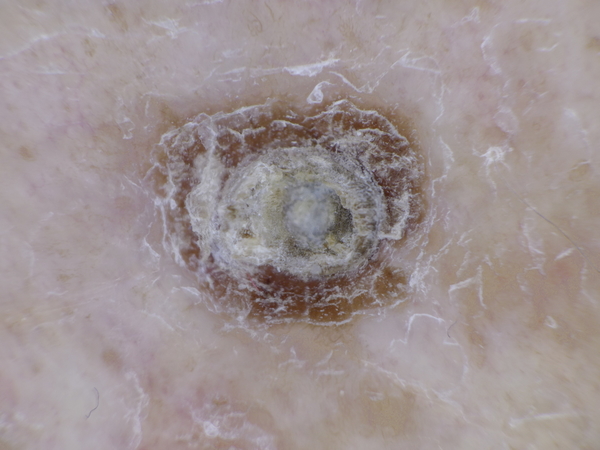}} 
\caption{Comparison between TS and ATS calibration method for calibrating the correctly classified and missclassified samples (more examples are provided in supplementary materials).}
\vspace{-0.3cm}
\label{calibration_TS_ATS}
 
\end{figure}

ATS improves calibration in order to have better referral system. To compare which method of TS or ATS can operate better as a referral system, we plot percentage of correctly classified samples that have confidence higher than the selected threshold versus percentage of misclassified samples that have confidence lower than that threshold. Figure \ref{fig_threshold} compares uncalibrated , calibrated with TS and ATS for different values of threshold which is changing in the interval of $[0,1]$ with step size 0.05. The Area Under Curve (AUC) is provided. It shows ATS by better calibration can improve the decision making system to refer less correctly classified cases to the specialist and do not miss the misclassified one by not referring them. Figure \ref{calibration_TS_ATS} gives some examples of confidence output of the system before and after calibration with TS and ATS methods. Comparatively, ATS can increase the gap between the confidence of correctly and misclassified samples more.

\section{Conclusion}
\label{conclusion}
Despite of dramatically improved accuracy of deep neural networks, they suffer from being overconfident. In this paper, we proposed ATS as a practical solution for calibrating DNNs. ATS is a Temperature  Scaling family method which tries to find the optimal $T$ value for rescaling the softmax layer and makes it more calibrated. ATS can find the optimal $T$ in the situation that the deep network is highly accurate, validation set is small or when it contains noisy labels where classic TS fails to find the best $T$. ATS is a post-processing approach that does not need to retrain the network for calibrating it which makes it an appropriate solution for calibrating already pretrained models and can be used in many applications as a practical solution.

{\small
\bibliographystyle{ieee}
\bibliography{egbib}

\begin{thebibliography}{10}\itemsep=-1pt

\bibitem{balan2015bayesian}
A.~K. Balan, V.~Rathod, K.~P. Murphy, and M.~Welling.
\newblock Bayesian dark knowledge.
\newblock In {\em Advances in Neural Information Processing Systems}, pages
  3438--3446, 2015.

\bibitem{bar2015chest}
Y.~Bar, I.~Diamant, L.~Wolf, S.~Lieberman, E.~Konen, and H.~Greenspan.
\newblock Chest pathology detection using deep learning with non-medical
  training.
\newblock In {\em ISBI}, pages 294--297. Citeseer, 2015.

\bibitem{bernardo2009bayesian}
J.~M. Bernardo and A.~F. Smith.
\newblock {\em Bayesian theory}, volume 405.
\newblock John Wiley \& Sons, 2009.

\bibitem{blundell2015weight}
C.~Blundell, J.~Cornebise, K.~Kavukcuoglu, and D.~Wierstra.
\newblock Weight uncertainty in neural networks.
\newblock {\em arXiv preprint arXiv:1505.05424}, 2015.

\bibitem{bojarski2016end}
M.~Bojarski, D.~Del~Testa, D.~Dworakowski, B.~Firner, B.~Flepp, P.~Goyal, L.~D.
  Jackel, M.~Monfort, U.~Muller, J.~Zhang, et~al.
\newblock End to end learning for self-driving cars.
\newblock {\em arXiv preprint arXiv:1604.07316}, 2016.

\bibitem{brier1950verification}
G.~W. Brier.
\newblock Verification of forecasts expressed in terms of probability.
\newblock {\em Monthey Weather Review}, 78(1):1--3, 1950.

\bibitem{chen2014stochastic}
T.~Chen, E.~Fox, and C.~Guestrin.
\newblock Stochastic gradient hamiltonian monte carlo.
\newblock In {\em International Conference on Machine Learning}, pages
  1683--1691, 2014.

\bibitem{DBLP:journals/corr/abs-1710-05006}
N.~C.~F. Codella, D.~Gutman, M.~E. Celebi, B.~Helba, M.~A. Marchetti, S.~W.
  Dusza, A.~Kalloo, K.~Liopyris, N.~K. Mishra, H.~Kittler, and A.~Halpern.
\newblock Skin lesion analysis toward melanoma detection: {A} challenge at the
  2017 international symposium on biomedical imaging (isbi), hosted by the
  international skin imaging collaboration {(ISIC)}.
\newblock {\em CoRR}, abs/1710.05006, 2017.

\bibitem{deng2009imagenet}
J.~Deng, W.~Dong, R.~Socher, L.-J. Li, K.~Li, and L.~Fei-Fei.
\newblock Imagenet: A large-scale hierarchical image database.
\newblock In {\em 2009 IEEE conference on computer vision and pattern
  recognition}, pages 248--255. Ieee, 2009.

\bibitem{friedman2001elements}
J.~Friedman, T.~Hastie, and R.~Tibshirani.
\newblock {\em The elements of statistical learning}, volume~1.
\newblock Springer series in statistics New York, 2001.

\bibitem{gal2016dropout}
Y.~Gal and Z.~Ghahramani.
\newblock Dropout as a bayesian approximation: Representing model uncertainty
  in deep learning.
\newblock In {\em international conference on machine learning}, pages
  1050--1059, 2016.

\bibitem{graves2013speech}
A.~Graves, A.-r. Mohamed, and G.~Hinton.
\newblock Speech recognition with deep recurrent neural networks.
\newblock In {\em Acoustics, speech and signal processing (icassp), 2013 ieee
  international conference on}, pages 6645--6649. IEEE, 2013.

\bibitem{guo2017calibration}
C.~Guo, G.~Pleiss, Y.~Sun, and K.~Q. Weinberger.
\newblock On calibration of modern neural networks.
\newblock {\em arXiv preprint arXiv:1706.04599}, 2017.

\bibitem{he2016deep}
K.~He, X.~Zhang, S.~Ren, and J.~Sun.
\newblock Deep residual learning for image recognition.
\newblock In {\em Proceedings of the IEEE conference on computer vision and
  pattern recognition}, pages 770--778, 2016.

\bibitem{hinton2015distilling}
G.~Hinton, O.~Vinyals, and J.~Dean.
\newblock Distilling the knowledge in a neural network.
\newblock {\em arXiv preprint arXiv:1503.02531}, 2015.

\bibitem{iandola2014densenet}
F.~Iandola, M.~Moskewicz, S.~Karayev, R.~Girshick, T.~Darrell, and K.~Keutzer.
\newblock Densenet: Implementing efficient convnet descriptor pyramids.
\newblock {\em arXiv preprint arXiv:1404.1869}, 2014.

\bibitem{jiang2011calibrating}
X.~Jiang, M.~Osl, J.~Kim, and L.~Ohno-Machado.
\newblock Calibrating predictive model estimates to support personalized
  medicine.
\newblock {\em Journal of the American Medical Informatics Association},
  19(2):263--274, 2011.

\bibitem{jin2017introspective}
L.~Jin, J.~Lazarow, and Z.~Tu.
\newblock Introspective classification with convolutional nets.
\newblock In {\em Advances in Neural Information Processing Systems}, pages
  823--833, 2017.

\bibitem{kirkpatrick2017overcoming}
J.~Kirkpatrick, R.~Pascanu, N.~Rabinowitz, J.~Veness, G.~Desjardins, A.~A.
  Rusu, K.~Milan, J.~Quan, T.~Ramalho, A.~Grabska-Barwinska, et~al.
\newblock Overcoming catastrophic forgetting in neural networks.
\newblock {\em Proceedings of the national academy of sciences},
  114(13):3521--3526, 2017.

\bibitem{krizhevsky2009learning}
A.~Krizhevsky and G.~Hinton.
\newblock Learning multiple layers of features from tiny images.
\newblock Technical report, Citeseer, 2009.

\bibitem{kuleshov2018accurate}
V.~Kuleshov, N.~Fenner, and S.~Ermon.
\newblock Accurate uncertainties for deep learning using calibrated regression.
\newblock {\em arXiv preprint arXiv:1807.00263}, 2018.

\bibitem{kumar2018trainable}
A.~Kumar, S.~Sarawagi, and U.~Jain.
\newblock Trainable calibration measures for neural networks from kernel mean
  embeddings.
\newblock In {\em International Conference on Machine Learning}, pages
  2810--2819, 2018.

\bibitem{lakshminarayanan2017simple}
B.~Lakshminarayanan, A.~Pritzel, and C.~Blundell.
\newblock Simple and scalable predictive uncertainty estimation using deep
  ensembles.
\newblock In {\em Advances in Neural Information Processing Systems}, pages
  6402--6413, 2017.

\bibitem{lecun1998gradient}
Y.~LeCun, L.~Bottou, Y.~Bengio, P.~Haffner, et~al.
\newblock Gradient-based learning applied to document recognition.
\newblock {\em Proceedings of the IEEE}, 86(11):2278--2324, 1998.

\bibitem{lecun1998mnist}
Y.~LeCun, C.~Cortes, and C.~Burges.
\newblock Mnist dataset, 1998.

\bibitem{liang2017enhancing}
S.~Liang, Y.~Li, and R.~Srikant.
\newblock Enhancing the reliability of out-of-distribution image detection in
  neural networks.
\newblock {\em arXiv preprint arXiv:1706.02690}, 2017.

\bibitem{louizos2016structured}
C.~Louizos and M.~Welling.
\newblock Structured and efficient variational deep learning with matrix
  gaussian posteriors.
\newblock In {\em International Conference on Machine Learning}, pages
  1708--1716, 2016.

\bibitem{louizos2017multiplicative}
C.~Louizos and M.~Welling.
\newblock Multiplicative normalizing flows for variational bayesian neural
  networks.
\newblock In {\em Proceedings of the 34th International Conference on Machine
  Learning-Volume 70}, pages 2218--2227. JMLR. org, 2017.

\bibitem{mackay1992bayesian}
D.~J. MacKay.
\newblock {\em Bayesian methods for adaptive models}.
\newblock PhD thesis, California Institute of Technology, 1992.

\bibitem{molchanov2017variational}
D.~Molchanov, A.~Ashukha, and D.~Vetrov.
\newblock Variational dropout sparsifies deep neural networks.
\newblock In {\em Proceedings of the 34th International Conference on Machine
  Learning-Volume 70}, pages 2498--2507. JMLR. org, 2017.

\bibitem{naeini2015obtaining}
M.~P. Naeini, G.~Cooper, and M.~Hauskrecht.
\newblock Obtaining well calibrated probabilities using bayesian binning.
\newblock In {\em Twenty-Ninth AAAI Conference on Artificial Intelligence},
  2015.

\bibitem{neal2012bayesian}
R.~M. Neal.
\newblock {\em Bayesian learning for neural networks}, volume 118.
\newblock Springer Science \& Business Media, 2012.

\bibitem{netzer2011reading}
Y.~Netzer, T.~Wang, A.~Coates, A.~Bissacco, B.~Wu, and A.~Y~Ng.
\newblock Reading digits in natural images with unsupervised feature learning.
\newblock {\em NIPS}, 01 2011.

\bibitem{neumann2018relaxed}
L.~Neumann, A.~Zisserman, and A.~Vedaldi.
\newblock Relaxed softmax: Efficient confidence auto-calibration for safe
  pedestrian detection.
\newblock In {\em Machine Learning for Intelligent Transportation Systems
  Workshop, NIPS}, 2018.

\bibitem{niculescu2005predicting}
A.~Niculescu-Mizil and R.~Caruana.
\newblock Predicting good probabilities with supervised learning.
\newblock In {\em Proceedings of the 22nd international conference on Machine
  learning}, pages 625--632. ACM, 2005.

\bibitem{platt1999probabilistic}
J.~Platt et~al.
\newblock Probabilistic outputs for support vector machines and comparisons to
  regularized likelihood methods.
\newblock {\em Advances in large margin classifiers}, 10(3):61--74, 1999.

\bibitem{ritter2018online}
H.~Ritter, A.~Botev, and D.~Barber.
\newblock Online structured laplace approximations for overcoming catastrophic
  forgetting.
\newblock In {\em Advances in Neural Information Processing Systems}, pages
  3742--3752, 2018.

\bibitem{ritter2018scalable}
H.~Ritter, A.~Botev, and D.~Barber.
\newblock A scalable laplace approximation for neural networks.
\newblock 2018.

\bibitem{simonyan2014very}
K.~Simonyan and A.~Zisserman.
\newblock Very deep convolutional networks for large-scale image recognition.
\newblock {\em arXiv preprint arXiv:1409.1556}, 2014.

\bibitem{tschandl2018ham10000}
P.~Tschandl, C.~Rosendahl, and H.~Kittler.
\newblock The ham10000 dataset, a large collection of multi-source
  dermatoscopic images of common pigmented skin lesions.
\newblock {\em Scientific data}, 5:180161, 2018.

\bibitem{wah2011caltech}
C.~Wah, S.~Branson, P.~Welinder, P.~Perona, and S.~Belongie.
\newblock The caltech-ucsd birds-200-2011 dataset.
\newblock 2011.

\bibitem{zadrozny2001obtaining}
B.~Zadrozny and C.~Elkan.
\newblock Obtaining calibrated probability estimates from decision trees and
  naive bayesian classifiers.
\newblock In {\em Icml}, volume~1, pages 609--616. Citeseer, 2001.

\bibitem{zadrozny2002transforming}
B.~Zadrozny and C.~Elkan.
\newblock Transforming classifier scores into accurate multiclass probability
  estimates.
\newblock In {\em Proceedings of the eighth ACM SIGKDD international conference
  on Knowledge discovery and data mining}, pages 694--699. ACM, 2002.

\bibitem{zagoruyko2016wide}
S.~Zagoruyko and N.~Komodakis.
\newblock Wide residual networks.
\newblock {\em arXiv preprint arXiv:1605.07146}, 2016.

\end{thebibliography}
}

\end{document}